\documentclass[letterpaper,10pt,conference]{ieeeconf}
\pdftrailerid{}
\usepackage{amsmath,amssymb}
\usepackage{booktabs}
\usepackage{graphicx}
\usepackage{multirow}
\usepackage{xcolor}
\usepackage{xspace}
\usepackage{cite}

\newcommand{\method}{InsightMap\xspace}
\newif\ifshowcitationpending
\showcitationpendingfalse 

\title{\LARGE\bfseries InsightMap: Structured Spatial Modeling for Embodied Multimodal Reasoning}

\author{Hongpei Zheng, Hujun Yin\\
University of Manchester}

\begin{document}
\bstctlcite{insightmap:reference-style}
\maketitle
\thispagestyle{empty}
\pagestyle{empty}

\begin{abstract}
Language-guided navigation requires connecting partial observations to a persistent spatial reference and learning how actions change that representation. We introduce \method, a framework that uses top-down maps as both explicit spatial memory and action-conditioned prediction targets. Historical views are linked to labeled map locations, and a shared multimodal backbone jointly learns navigation action prediction and post-action map generation. Map prediction provides auxiliary training supervision, while navigation inference decodes actions from the observed spatial context. An aligned RGB-D data pipeline supports a common interface for navigation, visual question answering, situated reasoning, and 3D grounding. On the validation-unseen splits of R2R-CE and RxR-CE, \method achieves success rates (SR) of 56.9\% and 54.9\%, respectively. Adding map-prediction supervision improves R2R-CE SR by 4.3 and success weighted by path length (SPL) by 3.2 percentage points. On static spatial tasks, \method achieves 103.7 CIDEr on ScanQA, 60.1\% exact-match accuracy on SQA3D, and 53.1\% grounding accuracy at 0.5 IoU on ScanRefer with detected object proposals. On Unitree Go2, it outperforms NaVid and NaVILA in hallway, lab, and office environments.
\end{abstract}

\section{Introduction}

Language-guided embodied tasks require robots to relate partial observations to a persistent spatial context. In vision-and-language navigation (VLN), this means connecting current views with previously visited places to follow an instruction. The resulting action--observation sequences also provide spatial transitions that can supervise policy learning.

Maps provide an explicit reference for organizing observation history. Topological, grid, multi-granularity, and annotated semantic maps aggregate spatial evidence and support navigation decisions~\cite{an2024etpnavevolvingtopologicalplanning, wang2023gridmmgridmemorymap, chen2022weaklysupervisedmultigranularitymaplearning, zhang2026mapnavnovelmemoryrepresentation}. However, action-label supervision alone does not explicitly train a policy to predict how its map changes after an action. This motivates using post-action maps as additional prediction targets for navigation learning.

\begin{figure*}[t]
    \centering
    \includegraphics[width=\textwidth,trim=135bp 71bp 86bp 59bp,clip]{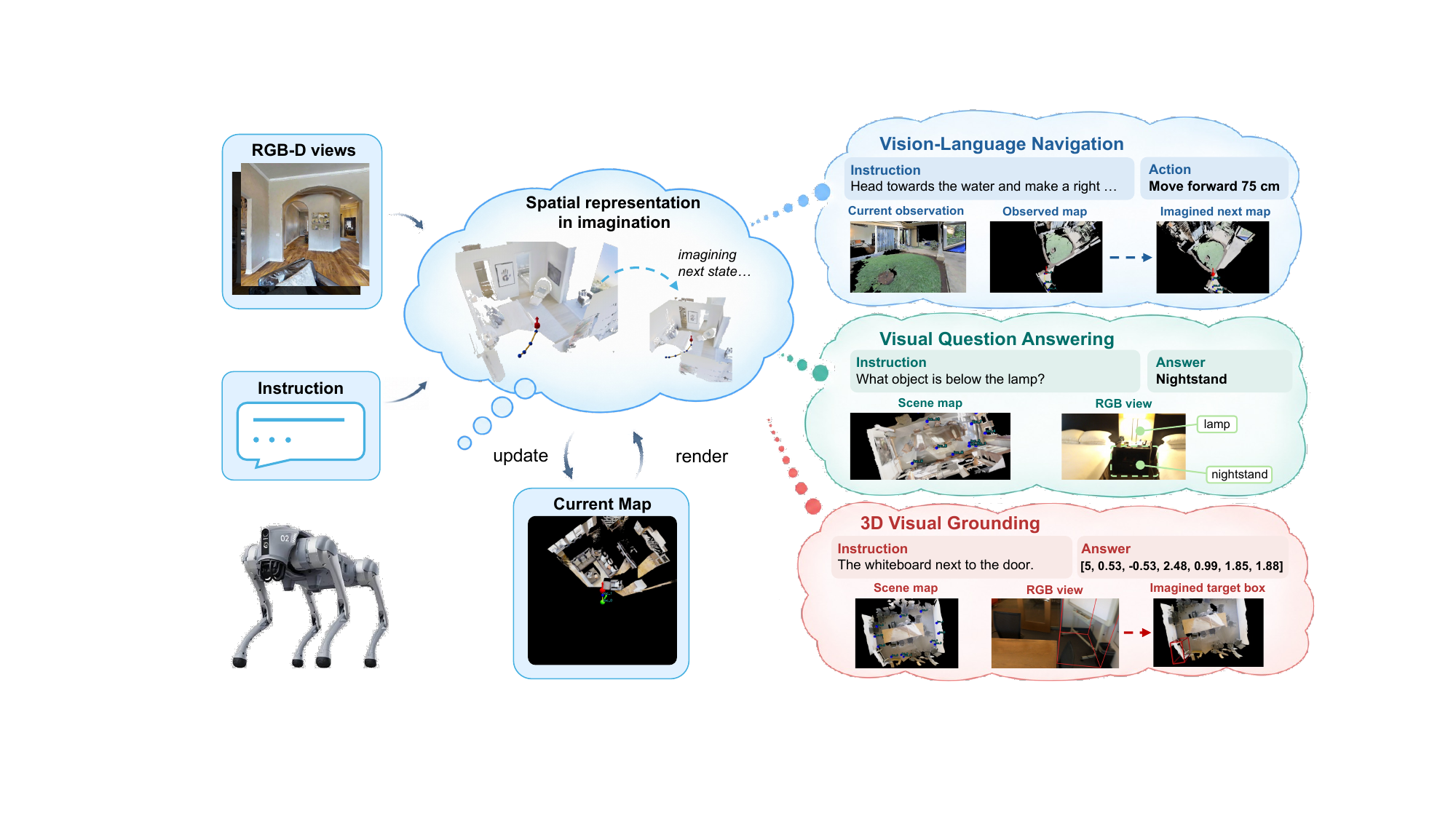}
    \caption{Overview of \method across vision-and-language navigation, visual question answering, and 3D visual grounding. An observed top-down map provides spatial context for all tasks. Navigation training pairs actions with post-action map targets. Question answering produces answer text, and grounding predicts a 3D box and a target-annotated map.}
    \label{fig:tasks}
\end{figure*}

Visual world models predict future observations, and World Action Models couple future visual prediction with action prediction~\cite{nvidia2025cosmosworldfoundationmodel, ye2026worldactionmodelszeroshot}. We apply this predictive principle to top-down spatial maps. Even in a static environment, ego-motion changes the map's local reference frame and new observations reveal additional geometry. Post-action map prediction therefore supplies a structured target for the evolution of accumulated spatial evidence.

We introduce \method, a framework that uses maps as both explicit spatial memory and action-conditioned prediction targets (Fig.~\ref{fig:tasks}). Historical first-person views are linked to labeled map locations, allowing visual observations and trajectory geometry to be interpreted together. Built on BAGEL~\cite{deng2025emergingpropertiesunifiedmultimodal}, the model jointly learns navigation action prediction and post-action map generation.

Question answering and 3D grounding also require associating objects and spatial relations across viewpoints. We incorporate these tasks into the same spatial interface to examine whether auxiliary spatial training benefits navigation. An aligned RGB-D data pipeline supplies navigation transitions, scene-level and situated question--answer pairs, and grounding examples with textual boxes and target-annotated maps.

We evaluate \method on R2R-CE and RxR-CE navigation, ScanQA and SQA3D question answering, and ScanRefer visual grounding. Adding map supervision improves R2R-CE success rate (SR) and success weighted by path length (SPL) by 4.3 and 3.2 percentage points, respectively. Adding historical location IDs to the map raises SR by 2.8 percentage points. The full task mixture achieves 56.9\% SR, compared with 49.2\% for navigation-only training. On Unitree Go2, \method outperforms NaVid and NaVILA in all three evaluated environments, with an SR gain of 45 percentage points over NaVILA in the office.

Our contributions are:
\begin{itemize}
    \item A map-based spatial interface that links historical views to labeled observation locations for navigation and static scene reasoning.
    \item A joint action-and-map training formulation that uses action-conditioned post-action map prediction as auxiliary spatial supervision for navigation.
    \item An aligned RGB-D data pipeline yielding 800K training examples, with benchmark and real-world evaluations that assess navigation, static spatial reasoning, and the effects of spatial supervision.
\end{itemize}

\section{Related Work}

\subsection{Embodied Navigation and Spatial Memory}
VLN models organize observation history through recurrent states and spatial maps. Recurrent VLN-BERT~\cite{Hong_2021_CVPR} carries a cross-modal state, ETPNav~\cite{an2024etpnavevolvingtopologicalplanning} builds an evolving topological map, and WS-MGMap~\cite{chen2022weaklysupervisedmultigranularitymaplearning} and GridMM~\cite{wang2023gridmmgridmemorymap} aggregate spatial features in metric and grid representations.

Recent systems use vision-language models to connect visual context with action selection. NaVid~\cite{zhang2024navidvideobasedvlmplans} predicts actions from video history, NaVILA~\cite{cheng2025navilaleggedrobotvisionlanguageaction} couples a high-level VLA with locomotion control, and InstructNav~\cite{long2024instructnavzeroshotgenericinstruction} uses language reasoning and value maps. MapNav~\cite{zhang2026mapnavnovelmemoryrepresentation} annotates an online semantic map with object labels to reduce reliance on historical frames. \method retains sampled historical views and explicitly links them to labeled map locations.

Spatial prediction also supports navigation learning and planning. BEVBert \cite{an2023bevbertmultimodalmappretraining} predicts semantic labels of masked map regions during pre-training. HNR \cite{wang2024lookaheadexplorationneuralradiance} predicts features of candidate locations for lookahead path evaluation. \method uses reconstructed maps as spatial memory and action-conditioned post-action map prediction as auxiliary training supervision.

\subsection{3D Question Answering and Grounding}
ScanQA~\cite{azuma2022scanqa3dquestionanswering}, SQA3D~\cite{ma2023sqa3dsituatedquestionanswering}, and ScanRefer~\cite{chen2020scanrefer3dobjectlocalization} evaluate scene-level question answering, situated reasoning, and language-guided 3D object localization, respectively.

Scene interfaces range from 3D features to object- and video-based representations. 3D-LLM~\cite{hong20233dllminjecting3dworld} lifts multi-view features into 3D, while LL3DA~\cite{Chen_2024_CVPR} operates on point clouds with text and visual prompts. Chat-Scene~\cite{huang2024chatscenebridging3dscene} uses object identifiers and object-centric embeddings, and 3D-LLaVA~\cite{deng20253dllavageneralist3dlmms} combines superpoint representations with mask prediction. Video-3D LLM~\cite{zheng2025video3dllmlearningpositionaware} incorporates global 3D position encodings into video features and selects views by maximum coverage.

GPT4Scene~\cite{qi2025gpt4sceneunderstand3dscenes} establishes global--local correspondence through a reconstructed BEV image and consistent object markers across views. \method uses camera-location labels to associate selected views with the scene map. For grounding, it couples frame-indexed 3D box prediction with generation of a target-annotated map; QA uses the shared spatial context with answer-text supervision.

\begin{figure*}[t]
    \centering
    \includegraphics[width=\textwidth,trim=58bp 70bp 24bp 40bp,clip]{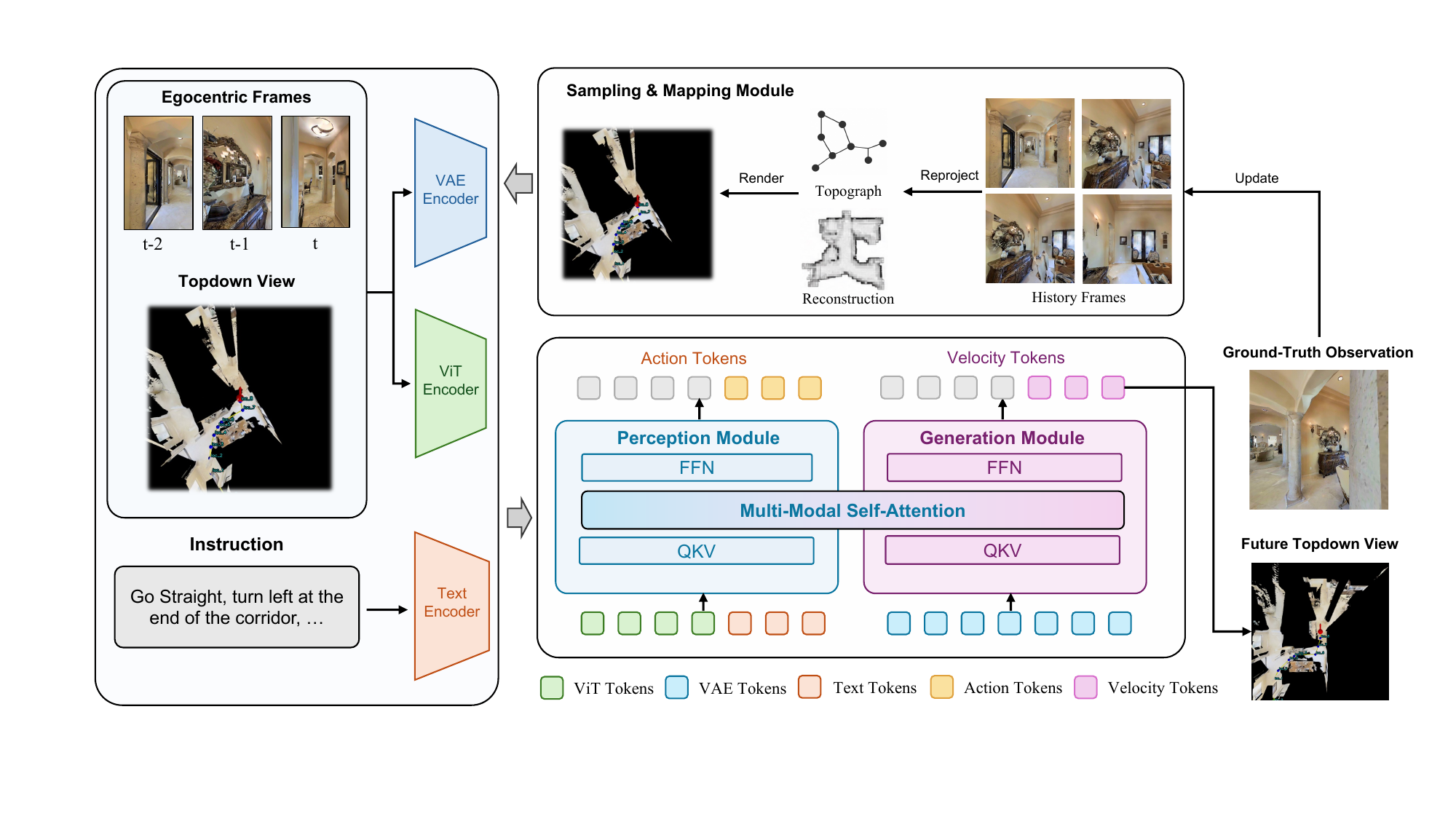}
    \caption{Architecture of \method. Labeled historical views, the current observation and map, and the instruction condition a shared understanding and generation backbone. The training targets are action text followed by an action-conditioned post-action map. }
    \label{fig:overview}
\end{figure*}

\subsection{Unified Understanding, Generation, and World Models}
BAGEL~\cite{deng2025emergingpropertiesunifiedmultimodal} unifies understanding and generation through a Mixture-of-Transformers backbone, trained with next-token prediction for text and rectified flow for images. \method adopts this architecture and these objectives, using spatial maps as visual targets aligned with action or grounding text.

Cosmos~\cite{nvidia2025cosmosworldfoundationmodel} formulates visual world modeling as predicting future observations conditioned on history and an action or instruction. DreamZero~\cite{ye2026worldactionmodelszeroshot} jointly generates future video and actions. \method applies this principle of coupling visual prediction with action learning to action-conditioned post-action maps as structured prediction targets.

\section{Method}
\label{sec:method}

\method couples spatial memory with structured world imagination (Fig.~\ref{fig:overview}): it anchors visual experience in an explicit map and learns to predict how this representation evolves after an action. Joint supervision of actions and their post-action maps connects embodied decision-making with prediction of spatial consequences. The same spatial interface supports question answering and grounding. Section~\ref{sec:data} describes the construction of aligned observations, maps, and targets.

\subsection{Spatial Memory}

To retain the spatial context of past observations, \method links historical views to labeled map locations. At navigation step $t$, let $\mathcal{H}_t=((o_i,\ell_i))_{i\in\mathcal{S}_t}$ denote the selected history in temporal order, where $\mathcal{S}_t$ indexes earlier macro states and $\ell_i$ labels the location of observation $o_i$. The context is $h_t=(q,\mathcal{H}_t,o_t,m_t)$: $q$ is the instruction, $o_t$ is the current observation, and $m_t$ is the reconstructed top-down map. The map contains the corresponding location labels, the connecting trajectory, and the current-pose marker. These view--location correspondences let visual history and trajectory geometry be interpreted together, preserving where evidence was acquired as the viewpoint changes.

\subsection{Structured World Imagination}

The map also provides a structured space in which to learn the spatial consequences of navigation. We formulate \emph{structured world imagination} as predicting the agent's post-action spatial representation, conditioned on the current context and action. The navigation training targets are an action $a_t$ and the map $m_{t+1}$ constructed after executing it. Their joint prediction is factorized as
\begin{equation}
    p(a_t,m_{t+1}\mid h_t)
    =p(a_t\mid h_t)\,p(m_{t+1}\mid h_t,a_t).
    \label{eq:navigation}
\end{equation}
The post-action map represents accumulated geometry in the agent's new local reference frame. Predicting it requires accounting for both ego-motion and newly observed geometry. During training, the context precedes the macro-action text and future map, represented as a continuous VAE latent. Ground-truth action tokens condition map generation, and the reference future map supplies supervision for the predicted spatial transition.

\emph{Navigation inference.} At each navigation step, \method decodes only the action text conditioned on $h_t$. After execution, new sensor observations update the observed map and visual history for the next decision. Future-map generation serves as auxiliary training supervision and is not performed during navigation evaluation.

\subsection{Unified Spatial Reasoning}

Spatial memory provides a common interface for relating language to visual evidence across tasks. For static scene reasoning, context $h$ contains the task text, selected views, and a scene map with corresponding observation-location labels. The text target $y$ is an answer or a frame-indexed 3D box representation. Grounding additionally supplies a target-annotated map $m^{*}$. Examples with both targets follow
\begin{equation}
    p(y,m^{*}\mid h)=p(y\mid h)\,p(m^{*}\mid h,y).
    \label{eq:joint}
\end{equation}
Equation~\ref{eq:navigation} is recovered with $h=h_t$, $y=a_t$, and $m^{*}=m_{t+1}$. The tasks share this interface while assigning different meanings to their outputs.

\emph{Visual grounding.} ScanRefer \cite{chen2020scanrefer3dobjectlocalization} pairs the spatial context with a referring expression. The text target is $[f,c_x,c_y,c_z,s_x,s_y,s_z]$, where $f$ identifies the first frame in which the referred object appears. The box center $(c_x,c_y,c_z)$ and size $(s_x,s_y,s_z)$ are expressed in that frame's camera coordinate system. The visual target $m^{*}$ adds the object's box to the map, providing a spatial annotation of the referred object. This target box is excluded from the input map.

\emph{Visual question answering.} ScanQA \cite{azuma2022scanqa3dquestionanswering} and SQA3D \cite{ma2023sqa3dsituatedquestionanswering} pair the spatial context with a question; SQA3D additionally includes the situation description. These tasks train only the answer distribution $p(y\mid h)$, with no target map.

\subsection{Joint Multimodal Learning}

The shared task interface uses a Mixture-of-Transformers (MoT) \cite{liang2025mixtureoftransformerssparsescalablearchitecture} backbone. First-person views and top-down maps are encoded through both ViT and VAE paths. The two visual paths provide complementary representations for action prediction and structured world imagination. ViT features capture high-level visual semantics that help relate observations to language instructions and support action prediction. VAE latents encode low-level visual detail in a representation that supports image reconstruction, providing the latent space for post-action map prediction. Text and ViT tokens follow the understanding branch, while VAE latents follow the generation branch. Shared self-attention allows both branches to integrate semantic context and visual detail for action prediction and map generation.

The understanding head predicts text tokens, and the generation head predicts rectified-flow velocities for target-map latents conditioned on the preceding context and text. Conditioning-image latents use the same generation branch but are not map-generation targets.

Task prompts and observations form packed text--image sequences, with supervision selecting the target text tokens and any target-map latents. Using the common context $h$ and text target $y$, the base text loss for one example at supervised positions $\mathcal{T}$ is
\begin{equation}
    \mathcal{L}_{\mathrm{CE}}=-\frac{1}{|\mathcal{T}|}
    \sum_{i\in\mathcal{T}}\log p_\theta(y_i\mid y_{<i},h).
    \label{eq:text-loss}
\end{equation}
For examples with a target map $m^{*}$, let $x_0$ be its clean VAE latent. We sample noise $x_1$ and a flow time $\tau\in[0,1]$, distinct from navigation time $t$, and construct
\begin{equation}
    x_\tau=(1-\tau)x_0+\tau x_1,\qquad v=x_1-x_0.
\end{equation}
The map-generation loss, conditioned on the preceding text target, is
\begin{equation}
    \mathcal{L}_{\mathrm{map}}=
    \frac{1}{|\mathcal{V}|}\sum_{j\in\mathcal{V}}
    \left\|\hat v_\theta(x_\tau,\tau,h,y)_j-v_j\right\|_2^2,
    \label{eq:map-loss}
\end{equation}
where $\mathcal{V}$ indexes the supervised target-map latent positions. For navigation, $y=a_t$ is the ground-truth action text; for grounding, it contains the target frame ID and box parameters. Writing $\widetilde{\mathcal{L}}_{\mathrm{CE}}$ for the length-reweighted text loss used in training, the combined objective is
\begin{equation}
    \mathcal{L}=\lambda_{\mathrm{CE}}\widetilde{\mathcal{L}}_{\mathrm{CE}}
    +\lambda_{\mathrm{map}}\mathcal{L}_{\mathrm{map}}.
\end{equation}
The MoT generation branch remains trainable during joint training, while the VAE is frozen. Navigation and grounding activate both losses. For QA, $\mathcal{V}$ is empty and we define $\mathcal{L}_{\mathrm{map}}=0$. Joint training therefore couples textual task prediction with explicit spatial supervision through the shared backbone.

\section{Spatial Data Construction}
\label{sec:data}

\begin{figure}[t]
    \centering
    \includegraphics[width=0.9\columnwidth]{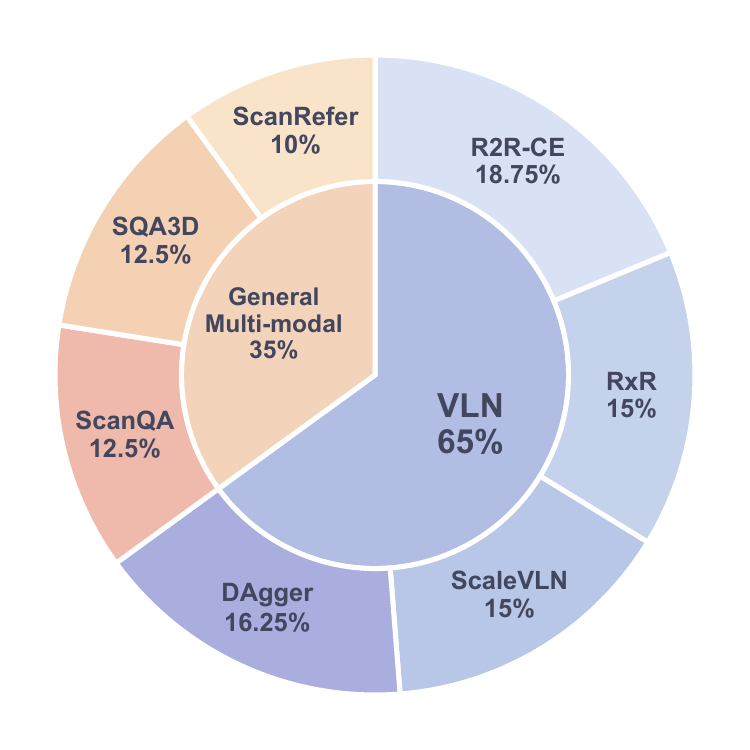}
    \caption{Composition of the 800K training examples. Navigation accounts for 65\%, and visual question answering and grounding for 35\%. All percentages refer to the full corpus.}
    \label{fig:training-data-mixture}
\end{figure}

We construct a training corpus of 800K examples that aligns visual observations, spatial context, and task supervision. Navigation contributes 65\% of the corpus, providing action-aligned map transitions for structured world imagination; visual question answering and grounding contribute the remaining 35\% (Fig.~\ref{fig:training-data-mixture}).

\subsection{Navigation Spatial Supervision}
Navigation data combine R2R-CE, RxR-CE, ScaleVLN~\cite{Wang_2023_ICCV}, and DAgger trajectories. Consecutive actions of the same type are grouped into macro actions. We pair each pre-action context with the action and the map reconstructed after execution, yielding aligned examples $(h_t,a_t,m_{t+1})$.

Trajectory replay in Habitat~\cite{savva2019habitatplatformembodiedai} supplies depth observations and poses for incremental truncated signed distance function (TSDF) reconstruction. Maps contain geometry observed up to each step, rendered over a fixed local region centered on the agent at its current floor, with image-up aligned to its heading. History views sampled at earlier macro states share location labels with the map, which also records the trajectory and current pose.

After one epoch of VLN-only training, we roll out the resulting policy and use expert guidance to construct corrective trajectories from erroneous navigation cases. These DAgger~\cite{ross2011reductionimitationlearningstructured} trajectories are aggregated with the original navigation data and provide both action and map supervision.

\subsection{Static Scene Construction}
For ScanQA~\cite{azuma2022scanqa3dquestionanswering}, SQA3D~\cite{ma2023sqa3dsituatedquestionanswering}, and ScanRefer~\cite{chen2020scanrefer3dobjectlocalization}, we construct full-scene maps from ScanNet geometry. Following Video-3D LLM~\cite{zheng2025video3dllmlearningpositionaware}, we greedily select RGB-D views that add the most scene-surface coverage and link them to indexed camera locations on the map.

We expand the VQA data through LLM-based question rewriting, conversion of object descriptions into question--answer pairs, and alternative situations for situated QA. Visual grounding data are augmented by rewriting referring descriptions. The resulting examples use the task targets defined in Section~\ref{sec:method}: answers for VQA, and boxes with target-annotated maps for grounding. Grounding input maps contain the scene and observation-location markers, excluding the target box.

\section{Experiments}
\label{sec:experiments}

\subsection{Experimental Setup}

\emph{Simulation Benchmark Setup.} We evaluate \method on the validation-unseen splits of R2R-CE \cite{krantz2020navgraphvisionandlanguagenavigationcontinuous} and RxR-CE \cite{ku2020roomacrossroommultilingualvisionandlanguagenavigation} in continuous Matterport3D environments through the Habitat simulator. For R2R-CE, we report navigation error (NE, in meters), oracle success rate (OS), success rate (SR), and success weighted by path length (SPL). OS, SR, and SPL are reported as percentages. For RxR-CE, we report NE, SR, SPL, and normalized dynamic time warping (nDTW) to additionally assess trajectory fidelity. Lower NE and higher values of the other metrics indicate better performance.

Throughout the tables, bold and underlined scores denote the best and second-best listed values, respectively. Dashes in metric columns indicate unavailable comparable results.

\emph{Real-World Evaluation Setup.} We deploy \method on a Unitree Go2 quadruped, using its built-in camera for visual observations and LiDAR-reconstructed point clouds for map construction. Model inference runs remotely on a server with a single NVIDIA H100 GPU. We evaluate navigation in hallway (easy), lab (medium), and office (hard) settings. \method, NaVid, and NaVILA are evaluated on the same robot platform, scenes, and navigation instructions.

\subsection{Implementation Details}

\emph{Training.} We build \method on BAGEL~\cite{deng2025emergingpropertiesunifiedmultimodal}. After one epoch of VLN-only training, the resulting policy is used to collect expert-guided DAgger corrections. Joint training uses the 800K-example corpus spanning navigation, visual question answering, and visual grounding described in Section~\ref{sec:data}. We fine-tune all learnable parameters except the VAE, including both MoT branches. Training uses AdamW with a learning rate of $2\times10^{-5}$ on eight NVIDIA H100 GPUs, accumulating gradients over four micro-steps for an effective global batch size of 64 examples per step. The text and map loss weights are $(\lambda_{\mathrm{CE}},\lambda_{\mathrm{map}})=(3,1)$.

\emph{Training inputs and actions.} Navigation examples contain the current RGB observation and up to ten history frames. ScanQA, SQA3D, and ScanRefer use 24, 24, and 16 views, respectively; all tasks also receive a map. The image transforms specify a maximum RGB size of 384 pixels and a target map resolution of $512\times512$, before encoder-specific alignment. Navigation maps cover approximately $16\,\mathrm{m}\times16\,\mathrm{m}$ around the agent by default. Constructed macro-action targets span $25$--$75$\,cm or $15$--$45^{\circ}$.

\subsection{Main navigation results}

Table~\ref{tab:main-navigation} compares \method with the listed navigation baselines. On RxR-CE, \method achieves NE 5.58, SR 54.9, SPL 46.1, and nDTW 62.8, the strongest values among the listed methods. Relative to NaVILA, SR and SPL improve by 5.6 and 2.1 percentage points, respectively, while nDTW increases by 4.0 points, indicating gains in task completion, path efficiency, and trajectory fidelity.

On R2R-CE, \method achieves the best listed OS and SR at 64.7 and 56.9, respectively. Its NE of 4.89 and SPL of 50.7 rank second, trailing ScaleVLN by 0.09\,m and 0.3 points. The results support competitive navigation and path efficiency within this comparison.

The comparison includes methods with different sensor inputs, training data, and waypoint predictors. Asterisks in Table~\ref{tab:main-navigation} denote source-designated simulator-trained waypoint predictors.

\begin{table}[t]
    \centering
    \caption{Navigation results on R2R-CE and RxR-CE val-unseen.}
    \label{tab:main-navigation}
    \small
    \setlength{\tabcolsep}{3pt}
    \begin{tabular*}{\columnwidth}{@{\extracolsep{\fill}}lcccc@{}}
        \toprule
        \multicolumn{5}{c}{\textbf{(a) R2R-CE}} \\
        \midrule
        Method & NE$\downarrow$ & OS$\uparrow$ & SR$\uparrow$ & SPL$\uparrow$ \\
        \midrule
        HPN+DN$^{*}$ \cite{krantz2021waypointmodelsinstructionguidednavigation} & 6.31 & 40.0 & 36.0 & 34.0 \\
        CMA$^{*}$ \cite{hong2022bridginggaplearningdiscrete} & 6.20 & 52.0 & 41.0 & 36.0 \\
        Sim2Sim$^{*}$ \cite{krantz2022sim2simtransfervisionandlanguagenavigation} & 6.07 & 52.0 & 43.0 & 36.0 \\
        GridMM$^{*}$ \cite{wang2023gridmmgridmemorymap} & 5.11 & 61.0 & 49.0 & 41.0 \\
        ScaleVLN$^{*}$ \cite{Wang_2023_ICCV} & \textbf{4.80} & -- & \underline{55.0} & \textbf{51.0} \\
        InstructNav \cite{long2024instructnavzeroshotgenericinstruction} & 6.89 & -- & 31.0 & 24.0 \\
        WS-MGMap \cite{chen2022weaklysupervisedmultigranularitymaplearning} & 6.28 & 47.6 & 38.9 & 34.3 \\
        Seq2Seq \cite{krantz2020navgraphvisionandlanguagenavigationcontinuous} & 7.77 & 37.0 & 25.0 & 22.0 \\
        NaVid \cite{zhang2024navidvideobasedvlmplans} & 5.47 & 49.1 & 37.4 & 35.9 \\
        MapNav \cite{zhang2026mapnavnovelmemoryrepresentation} & 4.93 & 53.0 & 39.7 & 37.2 \\
        NaVILA \cite{cheng2025navilaleggedrobotvisionlanguageaction} & 5.22 & \underline{62.5} & 54.0 & 49.0 \\
        \midrule
        \textbf{InsightMap} & \underline{4.89} & \textbf{64.7} & \textbf{56.9} & \underline{50.7} \\
        \midrule
        \multicolumn{5}{c}{\textbf{(b) RxR-CE}} \\
        \midrule
        Method & NE$\downarrow$ & SR$\uparrow$ & SPL$\uparrow$ & nDTW$\uparrow$ \\
        \midrule
        CMA$^{*}$ \cite{hong2022bridginggaplearningdiscrete} & 8.76 & 26.5 & 22.1 & 47.0 \\
        LAW \cite{raychaudhuri-etal-2021-language} & 10.90 & 8.0 & 8.0 & 38.0 \\
        Seq2Seq \cite{krantz2020navgraphvisionandlanguagenavigationcontinuous} & 12.10 & 13.9 & 11.9 & 30.8 \\
        NaVILA \cite{cheng2025navilaleggedrobotvisionlanguageaction} & 6.77 & \underline{49.3} & \underline{44.0} & \underline{58.8} \\
        UniNaVid \cite{zhang2025uninavidvideobasedvisionlanguageactionmodel} & \underline{6.24} & 48.7 & 40.9 & -- \\
        \midrule
        \textbf{InsightMap} & \textbf{5.58} & \textbf{54.9} & \textbf{46.1} & \textbf{62.8} \\
        \bottomrule
    \end{tabular*}
\end{table}

\begin{figure*}[t]
    \centering
    \includegraphics[width=\textwidth]{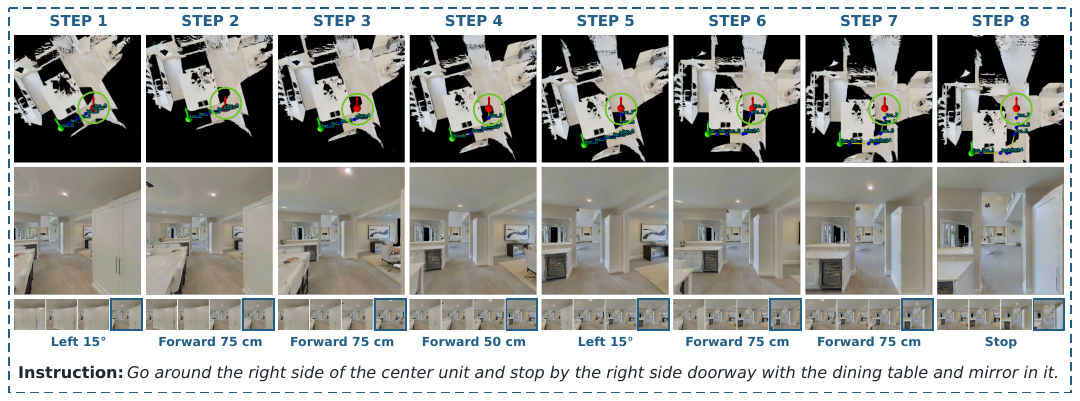}
    \caption{Recorded navigation trajectory of \method. The eight displayed steps pair top-down maps and current RGB views with history strips (past to current) and recorded actions. Green circles mark the agent, blue points indicate observation history, and yellow lines trace the path. The sequence ends with \texttt{STOP}.}
    \label{fig:vln-trajectory}
\end{figure*}

Figure~\ref{fig:vln-trajectory} shows the agent moving around the central unit toward the doorway specified by the instruction. As the viewpoint changes, the map retains the visited locations and traversed path, while the history strip preserves earlier visual observations. The sequence combines small heading adjustments with forward movements and ends with \texttt{STOP}. This example illustrates how the spatial representation connects changing local views with accumulated trajectory information throughout instruction following.

\subsection{Ablation studies}

We examine three factors on R2R-CE val-unseen: map-prediction supervision (Table~\ref{tab:ablation}), spatial map input and location-ID linking (Table~\ref{tab:ablation-spatial-input}), and auxiliary training tasks (Table~\ref{tab:ablation-task-mixture}).

\emph{Generation-module trainability and map supervision.} Table~\ref{tab:ablation} retains the full spatial input and conditioning-image VAE route in all variants. Unfreezing the generation module without map supervision (A1 to A2) improves SR and SPL by 2.3 and 2.4 percentage points. Adding map supervision at the same trainability (A2 to A3) yields further gains of 4.3 and 3.2 points, reaching 56.9 SR and 50.7 SPL. These results support map prediction as auxiliary supervision for navigation.

\begin{table}[t]
    \centering
    \caption{Ablation of generation-module trainability and map supervision on R2R-CE val-unseen.}
    \label{tab:ablation}
    \small
    \setlength{\tabcolsep}{3pt}
    \begin{tabular*}{\columnwidth}{@{\extracolsep{\fill}}lcccc@{}}
        \toprule
        Variant & Gen. module & Map loss & SR$\uparrow$ & SPL$\uparrow$ \\
        \midrule
        A1 & Frozen & Off & 50.3 & 45.1\\
        A2 & Trainable & Off & \underline{52.6} & \underline{47.5}\\
        \midrule
        \textbf{A3 (Full)} & Trainable & On & \textbf{56.9} & \textbf{50.7} \\
        \bottomrule
    \end{tabular*}
\end{table}

\emph{Spatial map input and location-ID linking.} Table~\ref{tab:ablation-spatial-input} keeps generation parameters trainable and future-map supervision unchanged. B1 uses a same-sized blank input map. B2 provides an unnumbered map, retaining geometry, location markers, trajectory, current pose, and RGB-frame labels. Relative to B1, B2 improves SR and SPL by 1.5 and 2.4 percentage points, showing the benefit of spatial context. Restoring historical location IDs on the map (B2 to B3) adds 2.8 and 0.8 points, supporting explicit links between historical views and map locations. Each input configuration applies during both training and evaluation, with the same map token budget and output resolution.

\begin{table}[t]
    \centering
    \caption{Ablation of map input and location IDs on R2R-CE val-unseen.}
    \label{tab:ablation-spatial-input}
    \small
    \setlength{\tabcolsep}{3pt}
    \begin{tabular*}{\columnwidth}{@{\extracolsep{\fill}}llccc@{}}
        \toprule
        Variant & Map input & Location IDs & SR$\uparrow$ & SPL$\uparrow$ \\
        \midrule
        B1 & Blank & No & 52.6 & 47.5\\
        B2 & Unnumbered map & No & \underline{54.1} & \underline{49.9}\\
        \midrule
        \textbf{B3 (Full)} & Full map & Yes & \textbf{56.9} & \textbf{50.7} \\
        \bottomrule
    \end{tabular*}
\end{table}


\emph{Task mixtures.} Table~\ref{tab:ablation-task-mixture} uses ScanRefer for Visual Grounding (VG) and both ScanQA and SQA3D for Visual Question Answering (VQA). C0 is the one-epoch VLN-only policy used to collect DAgger trajectories. C1 is retrained in a separate run using VLN and DAgger data and serves as the navigation-only reference. C1--C4 all include DAgger data. Adding VG (C2) or VQA (C3) improves SR/SPL over C1 by 2.5/2.4 or 5.9/4.7 percentage points, respectively. Combining both tasks (C4) achieves the best results across all four metrics, reaching 56.9 SR and 50.7 SPL, gains of 7.7 and 6.1 points over C1. Adding VG to VQA (C3 to C4) further improves SR/SPL by 1.8/1.4 points, supporting complementary benefits from the two auxiliary tasks.

\begin{table}[t]
    \centering
    \caption{Task-mixture ablations on R2R-CE val-unseen.}
    \label{tab:ablation-task-mixture}
    \small
    \setlength{\tabcolsep}{2pt}
    \begin{tabular*}{\columnwidth}{@{\extracolsep{\fill}}lcccccccc@{}}
        \toprule
        \multirow{2}{*}{Variant} & \multicolumn{3}{c}{Training tasks} & \multirow{2}{*}{DAgger} & \multicolumn{4}{c}{R2R-CE} \\
        \cmidrule(lr){2-4}\cmidrule(l){6-9}
        & VLN & VG & VQA & & NE$\downarrow$ & OS$\uparrow$ & SR$\uparrow$ & SPL$\uparrow$ \\
        \midrule
        C0 & $\checkmark$ & -- & -- & -- & 6.02 & 49.8 & 46.1 & 41.7\\
        C1 & $\checkmark$ & -- & -- & $\checkmark$ & 5.83 & 53.8 & 49.2 & 44.6\\
        C2 & $\checkmark$ & $\checkmark$ & -- & $\checkmark$ & 5.72 & 58.2 & 51.7 & 47.0\\
        C3 & $\checkmark$ & -- & $\checkmark$ & $\checkmark$ & \underline{5.09} & \underline{61.2} & \underline{55.1} & \underline{49.3}\\
        \midrule
        \textbf{C4} & $\checkmark$ & $\checkmark$ & $\checkmark$ & $\checkmark$ & \textbf{4.89} & \textbf{64.7} & \textbf{56.9} & \textbf{50.7}\\
        \bottomrule
    \end{tabular*}
\end{table}

\subsection{Static scene reasoning}

We evaluate static scene reasoning on ScanQA \cite{azuma2022scanqa3dquestionanswering} validation, SQA3D \cite{ma2023sqa3dsituatedquestionanswering} test, and ScanRefer \cite{chen2020scanrefer3dobjectlocalization} validation. Tables~\ref{tab:scanqa} and~\ref{tab:scanrefer} report task results alongside source-reported baselines.


For ScanQA, we report BLEU-4 (B-4), METEOR (M), ROUGE-L (R-L), and CIDEr (C). SQA3D uses exact-match accuracy (EM), reported as a percentage. Table~\ref{tab:scanqa} shows that \method achieves the best listed ScanQA validation results across all four metrics. Its CIDEr score of 103.7 exceeds NaVILA (16 frames) by 3.9 points. On SQA3D test, \method reaches 60.1\% EM, 5.53 percentage points above the best listed baseline, Chat-Scene. LL3DA is evaluated without test-time visual prompts. 

\begin{table}[t]
    \centering
    \caption{Question answering on ScanQA validation and SQA3D test.}
    \label{tab:scanqa}
    \small
    \setlength{\tabcolsep}{2pt}
    \begin{tabular*}{\columnwidth}{@{\extracolsep{\fill}}lccccc@{}}
        \toprule
        \multirow{2}{*}{Method} & \multicolumn{4}{c}{ScanQA} & SQA3D \\
        \cmidrule(lr){2-5}\cmidrule(l){6-6}
        & B-4$\uparrow$ & M$\uparrow$ & R-L$\uparrow$ & C$\uparrow$ & EM$\uparrow$ \\
        \midrule
        ScanQA \cite{azuma2022scanqa3dquestionanswering} & 10.08 & 13.14 & 33.33 & 64.86 & -- \\
        3D-LLM \cite{hong20233dllminjecting3dworld} & 12.0 & 14.5 & 35.7 & 69.4 & -- \\
        LL3DA \cite{Chen_2024_CVPR} & 13.53 & 15.88 & 37.31 & 76.79 & -- \\
        Chat-Scene \cite{huang2024chatscenebridging3dscene} & 14.31 & 18.00 & 41.56 & 87.70 & \underline{54.57} \\
        3D-LLaVA \cite{deng20253dllavageneralist3dlmms} & \underline{17.1} & 18.4 & 43.1 & 92.6 & 54.5 \\
        \midrule
        NaviLLM \cite{zheng2024learninggeneralistmodelembodied} & 12.0 & 15.4 & 38.4 & 75.9 & -- \\
        NaVILA \cite{cheng2025navilaleggedrobotvisionlanguageaction}  & 15.2 & \underline{19.6} & \underline{48.3} & \underline{99.8} & -- \\
        \midrule
        \textbf{InsightMap (Ours)} & \textbf{17.9} & \textbf{19.7} & \textbf{49.2} & \textbf{103.7} & \textbf{60.1} \\
        \bottomrule
    \end{tabular*}
\end{table}

Table~\ref{tab:scanrefer} reports overall ScanRefer box-localization accuracy (\%) at 3D intersection over union (IoU) thresholds of 0.25 and 0.5, denoted Acc@0.25 and Acc@0.5. \method directly generates the box center and size as text from the selected views and scene map, achieving raw accuracies of 43.8\% and 20.8\%. These exceed the raw results of VG-LLM~\cite{zheng2025learning} by 2.2 and 5.9 percentage points, respectively, showing stronger localization before proposal refinement within this comparison.

To refine these predictions, we match each predicted box to the candidate with the highest 3D IoU and use the matched proposal as the final box. The candidates are detected by Mask3D~\cite{schult2023mask3dmasktransformer3d} and provided by LEO~\cite{pmlr-v235-huang24ae}. Proposal matching raises Acc@0.25 and Acc@0.5 to 59.7\% and 53.1\%, gains of 15.9 and 32.3 percentage points over raw predictions. 

With proposal refinement, \method ranks second on both metrics among the listed methods, exceeding VG-LLM by 2.1 and 2.2 percentage points and trailing GPT4Scene by 2.9 and 3.9 points. These results show competitive grounding performance from the shared spatial interface, with detected proposals contributing substantially to the final accuracy.

\begin{table}[t]
    \centering
    \caption{ScanRefer validation results. Parentheses denote raw box predictions before proposal refinement.}
    \label{tab:scanrefer}
    \small
    \setlength{\tabcolsep}{3pt}
    \begin{tabular*}{\columnwidth}{@{\extracolsep{\fill}}lcc@{}}
        \toprule
        Method & Acc@0.25$\uparrow$ & Acc@0.5$\uparrow$ \\
        \midrule
        ScanRefer \cite{chen2020scanrefer3dobjectlocalization} & 41.19 & 27.40 \\
        Chat-Scene \cite{huang2024chatscenebridging3dscene} & 55.52 & 50.23 \\
        3D-LLaVA \cite{deng20253dllavageneralist3dlmms} & 51.2 & 40.6 \\
        Video-3D LLM \cite{zheng2025video3dllmlearningpositionaware} & 57.87 & 51.18 \\
        GPT4Scene \cite{qi2025gpt4sceneunderstand3dscenes} & \textbf{62.6} & \textbf{57.0} \\
        VG-LLM \cite{zheng2025learning} & 57.6 (41.6) & 50.9 (14.9) \\
        \midrule
        \textbf{InsightMap (Ours)} & \underline{59.7} (43.8) & \underline{53.1} (20.8) \\
        \bottomrule
    \end{tabular*}
\end{table}


\subsection{Real-world evaluation}
\label{sec:real-world}

\begin{figure}[t]
    \centering
    \includegraphics[width=\columnwidth,trim=40bp 20bp 0bp 32bp,clip]{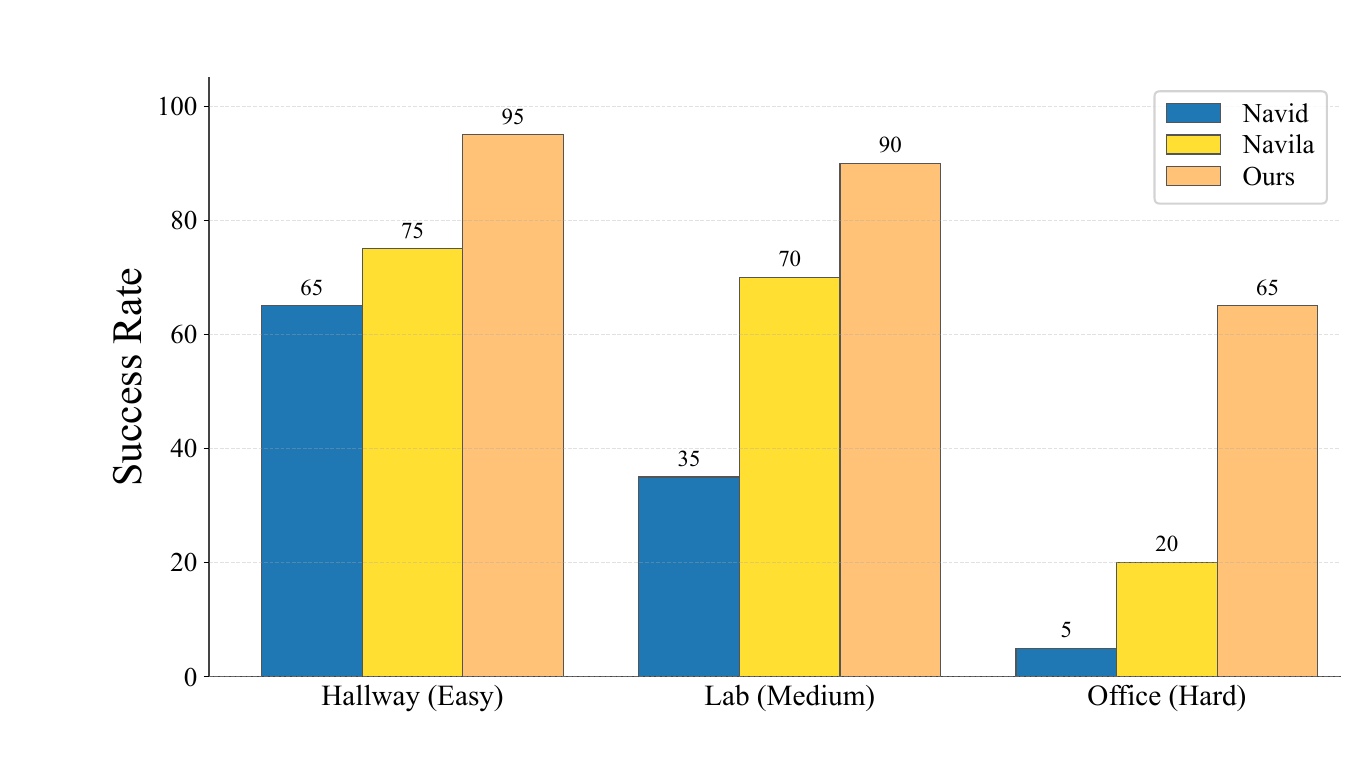}
    \caption{Real-world navigation success rates (\%) in hallway (easy), lab (medium), and office (hard) settings. All methods use the same Unitree Go2 platform, scenes, and navigation instructions.}
    \label{fig:real-world-metrics}
\end{figure}

We conducted 20 trials per scenario, measuring success rate (SR) in hallway, lab, and office settings. A trial is successful if the robot reaches within 2\,m of the goal within 50 steps. Figure~\ref{fig:real-world-metrics} summarizes the results. \method achieves 95\%, 90\%, and 65\%, exceeding NaVILA by 20, 20, and 45 percentage points, respectively. The largest improvement occurs in the office setting.

\begin{figure}[!htb]
    \centering
    \includegraphics[width=\columnwidth,trim=100bp 105bp 100bp 85bp,clip]{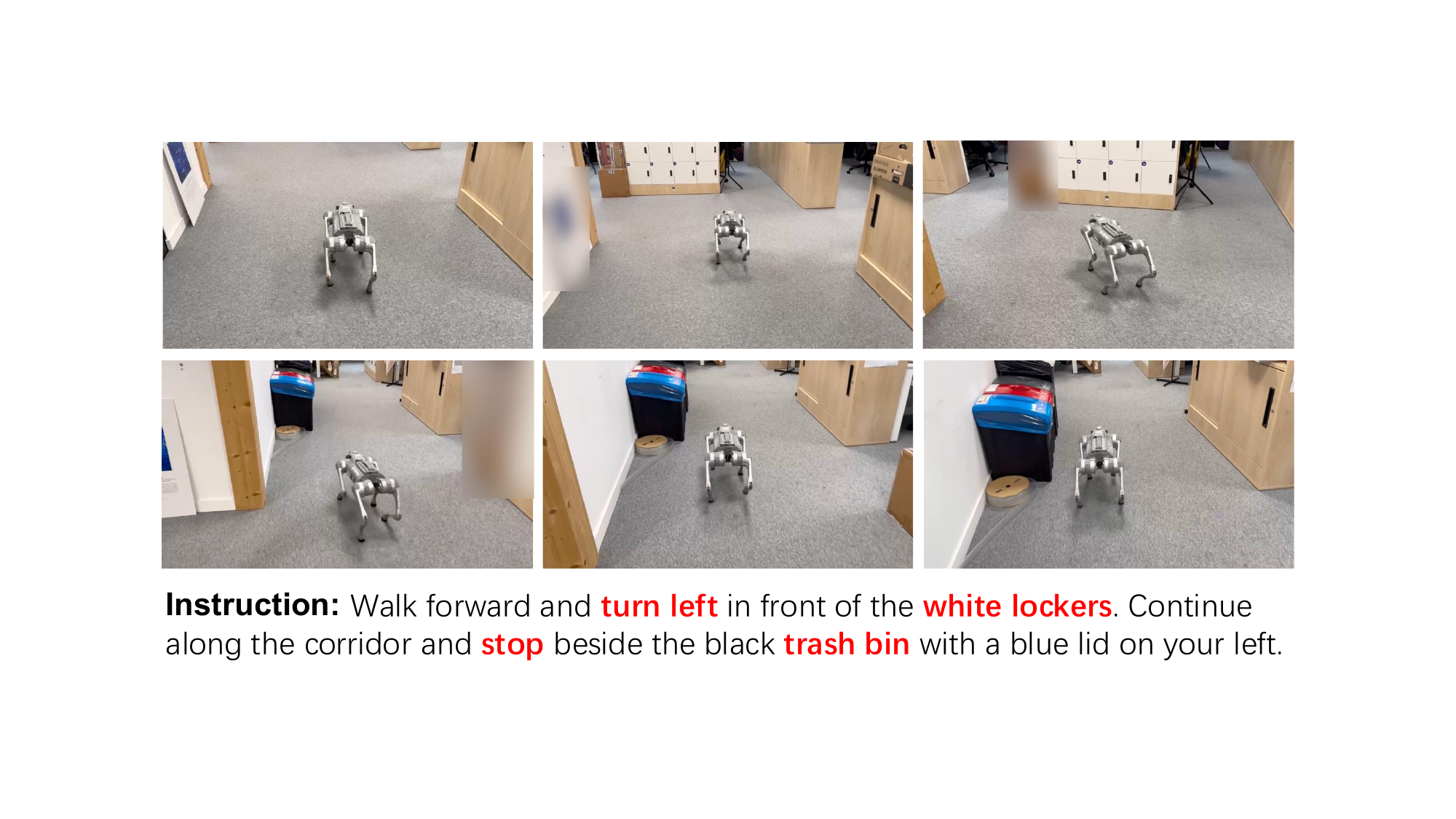}
    \caption{Real-world office navigation with \method on Unitree Go2. Third-person frames are ordered left to right, top to bottom.}
    \label{fig:real-world-office}
\end{figure}

The office example illustrates the need to link object descriptions with spatial relations over successive observations. The instruction combines a landmark-dependent left turn with a stopping location beside a second object. In Fig.~\ref{fig:real-world-office}, the robot approaches the white lockers, changes direction, and proceeds toward the black trash bin with a blue lid. This sequence provides a concrete example of carrying out a multi-stage instruction on the physical platform. Together with the success-rate comparison, it supports the applicability of the shared spatial interface to real-world navigation.

\section{Conclusion}

We presented \method, a framework that links historical views to map locations and jointly trains navigation action prediction and action-conditioned map generation. On R2R-CE val-unseen, adding map supervision improves SR and SPL by 4.3 and 3.2 percentage points. The navigation ablations also show higher SR with explicit view--location links and the full task mixture. Static scene evaluation yields 103.7 CIDEr on ScanQA, 60.1\% exact-match accuracy on SQA3D, and 53.1\% Acc@0.5 on ScanRefer with detected object proposals. Unitree Go2 trials show higher SR than NaVid and NaVILA in all three evaluated environments. These findings establish the practical value of organizing observation history and learning spatial transitions through a shared map representation.

\bibliographystyle{IEEEtran}
\bibliography{references}

\end{document}